\def\year{2017}\relax
\documentclass[letterpaper]{article} 
\usepackage{aaai17}  
\usepackage{times}  
\usepackage{helvet}  
\usepackage{courier}  
\usepackage{url}  
\usepackage{graphicx}  
\usepackage{wasysym}
\usepackage{multirow}
\usepackage{subfigure}
\usepackage{enumitem}
\begin{document}
%
\title{Robot-Initiated Specification Repair through Grounded Language Interaction}
\author{Adrian Boteanu$^{1}$, Jacob Arkin$^{2}$, Siddharth Patki$^{2}$, Thomas Howard$^{2}$ and Hadas Kress-Gazit$^{1}$
\thanks{$^{1}$Adrian Boteanu \texttt{(ab2633@cornell.edu)} and Hadas Kress-Gazit \texttt{(hadaskg@cornell.edu)} are with the Sibley School of Mechanical and Aerospace Engineering, Cornell University, Ithaca, NY. 
}
\thanks{$^{2}$Jacob Arkin \texttt{(j.arkin@rochester.edu)}, Siddharth Patki \texttt{(spatki@ur.rochester.edu)} and Thomas Howard \texttt{(thoward@cs.rochester.edu)} are with the Hajim School of Engineering and Applied Sciences, University of Rochester, Rochester, NY.}
\thanks{This work was supported by NSF NRI award 1427030.}}

\maketitle
\begin{abstract}

Robots are required to execute increasingly complex instructions in dynamic environments, which can lead to a disconnect between the user's intent and the robot's representation of the instructions. In this paper we present a natural language instruction grounding framework which uses formal synthesis to enable the robot to identify necessary environment assumptions for the task to be successful. These assumptions are then expressed via natural language questions referencing objects in the environment. The user is prompted to confirm or reject the assumption. We demonstrate our approach on two tabletop pick-and-place tasks.

\end{abstract}

\section{INTRODUCTION}

\begin{figure*}[tb]
\begin{center}
\includegraphics[width=0.7\textwidth]{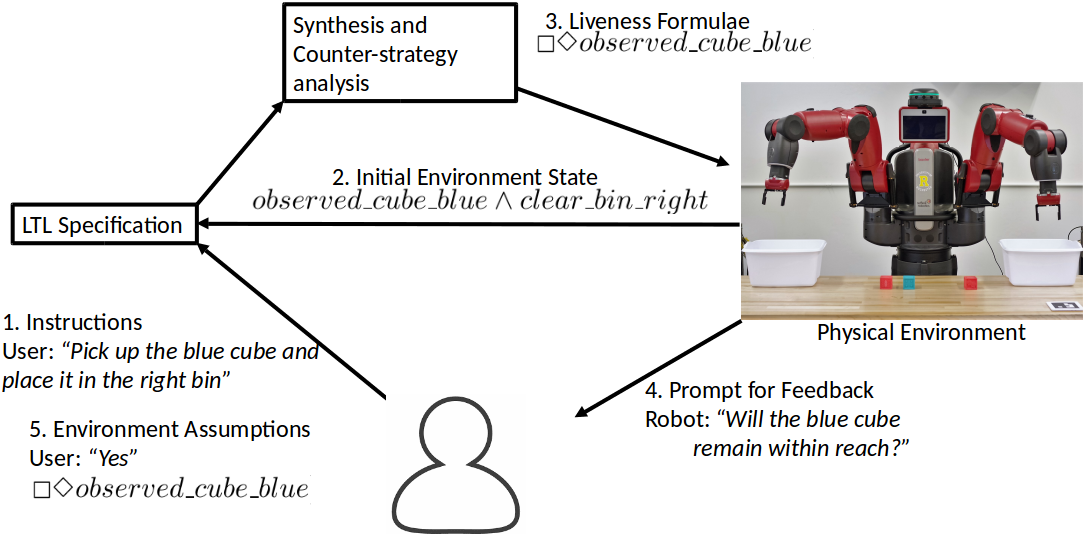}
\caption{User interaction diagram showing main steps of our approach. Completing grounded specifications with environment assumptions is an iterative process which begins with a user command, which the system grounds to an LTL specification. The specification is then analyzed and assumptions are derived from it. The user is then prompted to confirm or reject these assumptions.
\label{fig:interaction}}
\end{center}
\end{figure*}

Establishing links between symbols such as language and their physical manifestation, a process known as symbol grounding \cite{harnad1990symbol}, is a central problem in natural language understanding for human-robot interaction (HRI) \cite{stubbs2007autonomy,lemaignan2012grounding}. Grounded language enables robots to interpret and respond to instructions, allowing information exchange about the physical world between humans and robots. The outcome of grounding affects the robot autonomy \cite{goodrich2007human}. 

As robots become more autonomous and perform increasingly complex tasks, more descriptive groundings are needed in order to represent these tasks. This representation problem is further complicated by possibly incomplete or ambiguous natural language instructions (NLI), for example excluding elements that users consider self-evident. This creates the potential for the robot to incorrectly interpreting user intent. 
In particular, complex instructions consisting of multiple interdependent steps propagate these errors: the sorting instruction in Figure \ref{fig:interaction} implies performing a ``pick up'' action on ``the blue cube,'' followed by a ``place'' action ``in the right bin.'' The success of the drop action depends on first correctly executing the pick-up.

One existing method of grounding NLI employs probabilistic graphical models, which interpret single instructions (e.g. ``pick up the crate'') by mapping phrases to objects and actions \cite{tellex2011understanding}. An evolution of this approach, the Distributed Corespondence Graph (DCG), learns symbolic constraints to limit the search space of the grounding problem, allowing the use of large symbolic representations for grounding \cite{howard2014natural}. A limitation of these models is that the task plan produced through grounding is not assessed for correctness, potentially leading to inconsistencies and errors that propagate along multiple execution steps.

Physical groundings have been complemented with verifiable logical formluae in order to enable robots to execute complex instructions of conditionally dependent steps \cite{boteanu2016model}. The Verifiable-DCG model (V-DCG) extends the DCG symbolic representation to include both physical objects and Linear Temporal Logic (LTL) formulae. Grounding instructions using V-DCG generates formal LTL specifications which are then synthesized into verified controllers. Successful synthesis guarantees that the high-level symbolic task is achievable. Conversely, synthesis failures reveal logical conflicts in the specification that make it impossible to execute in all or some environments defined in the specification.

If synthesis is unsuccessful (i.e. the specification is \textit{unsynthesizable}), the robot player cannot achieve its task. Unsynthesizable specifications are either \textit{unsatisfiable} or \textit{unrealizable}. 
The robot is unable to follow \textit{unsatisfiable} specifications in any environment; for example in the sorting environment shown in Figure \ref{fig:env2}, an unsatisfiable specification would require the robot to both pick up a red block with its left gripper and simultaneously to never pick up with any of its grippers. Unrealizable specifications only allow the robot to respond to environment changes for a finite number of steps, after which the robot is unable to continue executing its task. Unrealizable specifications imply that there exists at least an \textit{admissible} environment in which the robot is unable to achieve its task \cite{raman2013explaining}. An admissible environment is a state permitted by the specification which can be reached in a finite number of steps starting from one of the possible initial environment states.

The V-DCG model is able to identify unsatisfiable specifications obtained from grounding NLI, highlighting errors in the task description (e.g. requesting a drop action without a corresponding pickup).
In this paper we contribute the following by building on the V-DCG model:
\begin{itemize}[nolistsep,noitemsep]
\item We expand the V-DCG grounding symbol hierarchy to allow richer representations;

\item We provide a method for situated natural language interaction between the robot and the user, in which the robot can respond to NLI from the user by requesting approval for environment assumptions. Our approach has the following benefits: (1) the user can provide assumptions without expert knowledge about the robot's capabilities; (2) the system is proactive and initiates user interaction as needed, guiding the user in providing necessary assumptions; (3) the final specification more accurately represents the execution environment, since its assumptions are established through situated interaction.

\item We demonstrate our model on a significantly larger instruction corpus (two tasks and 60 instructions, from one task and six instructions \cite{boteanu2016model}).

\end{itemize}

The core contribution of our paper, obtaining environment assumptions through user interaction, is an iterative process (Figure \ref{fig:interaction}):

\begin{enumerate}
\item The user states an instruction which is grounded to a LTL specification using a pre-trained model. The system then attempts to synthesize the LTL specification into a controller. If the specification is synthesizable, execution begins. If it is unsatisfiable, existing work can inform the user about possible causes \cite{raman2013sorry,boteanu2016model}. Our contribution targets unrealizable outcomes, in which some admissible environments can impede the robot, thus the following assumes the initial specification is unrealizable;

\item We supplement the specification with the environment's initial state as perceived by the robot, and reattempt synthesis;

\item If the specification is again unrealizable, we extract the collection of environment moves that will make the robot fail, which are known as \textit{counter-strategy}. An unrealizable specification can be executed for a finite number of steps until the robot is unable to respond to environment changes. These are worst-case scenarios which are possible given the current specification. While for unrealizable specifications there may also be environments in which the robot can successfully execute its goals indefinitely, one advantage of our approach to use formal verification in generating controllers is that we obtain a worst-case analysis of all possible execution sequences. We use existing work in counter-strategy analysis to generate environment assumption formulae that would, if added to the specification, produce a synthesizable specification \cite{alur2013counter};

\item We use the proposition evaluation results and the assumption formulae derived from the counter-strategy to generate prompts, which the user can either accept or reject;

\item If the user answers a prompt affirmatively, we add the corresponding formulae to the specification. We then attempt to re-synthesize the specification, continuing steps 3, 4 and 5 until synthesis succeeds or the user denies all prompts generated by the system. Once synthesis has succeeded, the robot executes the task using the resulting reactive controller.

\end{enumerate}

The rest of the paper is structured as follows: we first review relevant literature, and summarize the LTL formalism and the V-DCG model.
We then describe our contribution in generating environment assumptions and feedback prompts.
Finally, we evaluate our grounding model on two tasks: a binary cube sorting task and a cube stacking task, for which we crowdsourced and annotated an instruction corpus.

\section{RELATED WORK}\label{sec:related_work}

Natural language grounding and natural language interaction have been studied in HRI from a broad range of perspectives ranging from psychological motivations to enabling robot autonomy \cite{roy2005grounding}. To the best of our knowledge, our work is the first to incorporate stochastic language grounding, robot-driven language interaction for task repair, and generating language from formal synthesis results. We will now summarize existing work in these areas.

One approach to grounding NLI uses probabilistic models which map the meaning of parsed language to motion-planned trajectories, regions in physical space and objects (Generalized Grounding Graph, $G^3$) \cite{tellex2011understanding}. The DCG model extends the $G^3$ model by defining symbolic constraints which partition physical space prior to planning \cite{howard2014natural}. A central feature of the DCG model is that, in addition to physical groundings such as objects and trajectories, the grounding can use arbitrary symbolic representations. For example, the original model used spatial symbols to describe concepts such as \textit{near}, \textit{right} and \textit{left}, while recent contributions have introduced abstract groupings such as rows of objects \cite{Paul2016Abstract}. In these approaches, individual grounding results are evaluated together over instruction sequences, which can lead to inconsistencies between different instructions as a result of the generative nature of these models. The V-DCG model introduces logical groundings which allow instruction sequences to be holistically verified \cite{boteanu2016model}. 


Human-robot interaction for grounding has been shown in robot learning from demonstration \cite{argall2009survey}. When interpreting instructions, the robot would recursively prompt the user to clarify unknown groundings \cite{lauriar2001training,chao2011towards}. 
Stochastic grounding models have been used to generate salient object references to allow the robot to request help from users, enabling recovery from execution exceptions that the robot cannot address by itself \cite{tellex2014asking}. 
Voice prompts have also been used to enable users to help robots adapt known tasks to new environments by requesting the user to verify substitutes proposed by the robot \cite{boteanu2015towards}.

Existing work that leveraged formal synthesis to generate verbal feedback \cite{raman2013sorry} relied on manually-defined groundings for actions, whereas our model learns these symbols. Other work in formally representing robot instructions uses Combinatorial Categorical Grammars to infer logical representations of navigation instructions \cite{matuszek2013learning}.

\section{PRELIMINARIES}\label{sec:prelim}

\subsection{Linear Temporal Logic Overview}\label{sec:prelim_ltl}
Linear Temporal Logic (LTL) is defined over propositions and formulae taking Boolean values over an infinite discrete time series. Propositions are atomic variables which hold a binary truth value at each time step and offer discrete abstractions: \textit{Sensor} propositions, abstracting the \textit{environment}, $\{x_i | i=1..n\}$, become active when a perception condition is met, for example an object is recognized; \textit{Action} propositions, abstracting the \textit{robot}, $\{\alpha_i | i=1..m\}$, command actuation primitives such as navigation and grasping. In addition, the system can represent \textit{memory} propositions, which correspond to internal states.
Formulae, $\phi$, are formed by applying Boolean ($\neg, \land, \lor$) and temporal operators (next, $\ocircle$, until, $U$, together with the derived symbols eventually, $\Diamond$ and always, $\Box$) to proposition and other formulae. The temporal operators have the following semantics: $\ocircle \phi$ -- at the next time step the $\phi$ will be True; $\psi U \phi$ -- at some future time step $\phi$ will be True, and $\psi$ must be true until then; $\Diamond \phi$ -- at some future time step, $\phi$ will be true; $\Box \phi$ -- $\phi$ is True for every time step.
For example, $\Box \Diamond \phi$ means that the formula will repeatedly become \textit{True}; $\Box ( x_i \to \alpha_j )$ expresses that at all time steps the logical implication holds.

We use the GR(1) fragment of LTL since it offers tractable synthesis \cite{piterman2006synthesis}. GR(1) formulae follow an assume-guarantee structure describing a two-player game in which the robot player responds through actions to changes in sensors controlled by an adversarial environment player; the robot's behavior is thus reactive to the environment. 
GR(1) specifications abstract tasks as a two-player game: the \textit{robot} and the \textit{environment} in which the robot operates, both operating under formal assumptions described through LTL formulae \cite{kress2008translating}.
The robot player has control over \textit{action propositions}, which interface with actuation, while the environment player controls \textit{sensor} propositions, which correspond to the robot's perception. Both players' behaviors are restricted through \textit{assumptions}. Assumptions restricting the behavior of either player are considered \textit{safeties}, while \textit{liveness} are assumptions that express goals. For the example in Figure \ref{fig:interaction}, an environment assumption states that a block will eventually appear in the environment, and the robot can only perform a pick up action if it senses a block. The specification defines an initial state for the game, $\phi_i$.
If successful, synthesis produces a controller that guarantees the robot's behavior under environment assumptions \cite{kress2009temporal}.

\subsection{V-DCG Overview}\label{sec:vdcg_prelim}

\begin{figure}
\begin{center}
\includegraphics[width=0.3\textwidth]{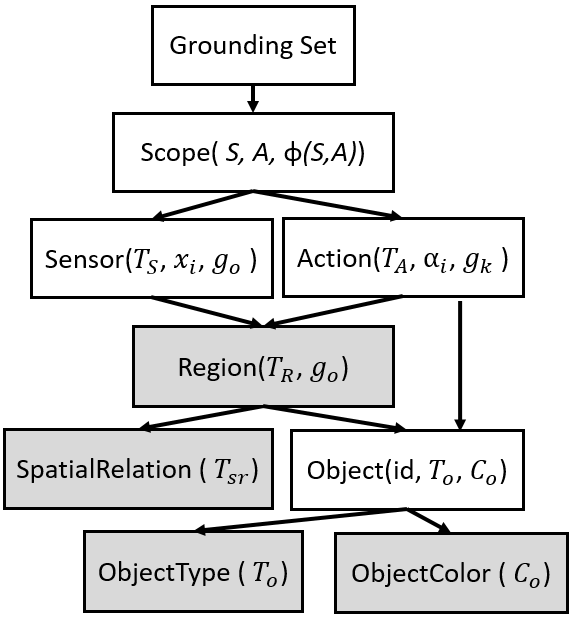}
\caption{We expand the symbolic hierarchy of the V-DCG model by adding \textit{ObjectType}, \textit{ObjectColor}, \textit{SpatialRelation} and \textit{Region} symbols (in grey).
\label{fig:symbols}}
\end{center}
\end{figure}

The V-DCG grounding model generates LTL specifications by grounding natural language using the following symbol hierarchy:

\begin{itemize}[nolistsep,noitemsep]
\item \textit{Object}( \textit{id}, $T_o$, $C_o$ ), where \textit{id} is a unique object identifier produced by the perception system; $T_o$ is a type from a known set of object types, for example \textit{\{cube, bin, robot gripper\}}; $C_o$ is a color from a given set, for example \textit{\{blue, red, green\}}. These properties can be easily extended by adding different attributes and possible values for each attribute;

\item \textit{Sensor}( $T_S$, $x_i$, $g_o$ ), where $T_S$ denotes a sensor type which indicates which robot perception primitive should be invoked when evaluating this proposition; $x_i$ is the proposition as it is used in the LTL specification; $g_o$ are grounded \textit{Object} types associated with this sensor;

\item \textit{Action}( $T_A$, $\alpha_i$, $g_k$ ), where $T_A$ indicates the robot's actuation primitive to be invoked when activating the proposition corresponding to this action; $\alpha_i$ is the proposition used in the LTL specification; $g_k$ are grounded objects which are used by the actuation function $T_A$;

\item \textit{Scope}( $S$, $A$, $\phi(S,A)$ ) combines sensors, $S$, and action $A$. Depending on the types of sensors and actions, different formulae $\phi(S,A)$ are produced. This mapping is defined as part of the grounding model;

\item \textit{Grounding Set} is the union of all other grounding symbols inferred from a parsed instruction; instructions are parsed using a CYK parser \cite{gonzalez1978syntactic} and a production grammar to annotate the sentence with Penn TreeBank tags \cite{marcus1994penn}. The specification conjuncts all LTL formulae $\phi(S,A)$ present in \textit{Scopes} of the \textit{Grounding Set}.

\end{itemize}

The V-DCG model assumes that the environment \textit{safety} formulae defined during model training hold for new grounding environments. If the environment exhibits behaviors not captured by these formulae and violates the specification, execution is halted. 
In this paper we enable specifications generated from grounding NLI in new environments to be interactively supplemented by the user, such that any necessary assumptions that were not captured during model training are added before execution.

\section{SYMBOLIC REPRESENTATION}\label{sec:symbols}

\begin{figure}
\begin{center}
\includegraphics[width=0.35\textwidth]{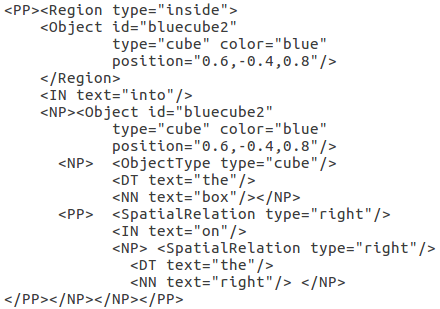}
\caption{Annotation examples showing symbols introduced in this work which expand the V-DCG representation: \textit{Object Type}, \textit{Spatial Relation} and \textit{Region}.} \label{fig:annotation}
\end{center}
\end{figure}

We extend the V-DCG symbol hierarchy with new symbols (Figure \ref{fig:symbols}). Firstly, we enable the model to better learn from ambiguous object references by leveraging spatial information relative to the robot. For example, an ambiguous phrase such as ``the box'' in an environment with multiple cubes (Figure \ref{fig:annotation}) can be clarified by the spatial reference ``on the right''. To do so, we added the symbols: 

\begin{itemize}[nolistsep,noitemsep]
\item \textit{ObjectType}, which have the same types as \textit{Objects}, $T_o$;
 
\item \textit{ObjectColor}, with $C_o$ the same as for \textit{Objects};
 
\item \textit{SpatialRelation}$\in$\textit{\{center, left, right, above\}}.
\end{itemize}

Secondly, in order to correctly associate actions and objects, we introduce \textit{Region} symbols. \textit{Regions} associate \textit{Objects} with prepositions (e.g. ``\textit{on} the red block''), thus restricting the space of possible actions that can be performed on that object. 

Figure \ref{fig:annotation} shows an example of how these symbols are used to ground the phrase \textit{``into the box on the right''}. First, an \textit{ObjectType} symbol of type \textit{cube} is associated with the Noun Phrase (NP) \textit{the box}, then \textit{SpatialRelation} symbols are associated with the prepositional phrase \textit{on the right} and the noun phrase \textit{the right}. These symbols are combined in a NP to identify an \textit{Object} of type \textit{cube}, with known color and position. A \textit{Region} is associated with the preposition ``into'' and is used to describe the Prepositional Phrase (PP) representing the phrase.

The final LTL formula associated with a scope is constructed by mapping proposition types and their objects to formula templates which establish conditional dependence between the propositions and are determined by the task type. We will give two examples, in which \textit{observed\_cube\_blue} and \textit{understack\_cube\_red} are sensor propositions, \textit{pickup\_right} is an action proposition, and \textit{right\_gripper} is a memory proposition:

The phrase \textit{``Pick up the blue cube with your right hand''} for the sorting task is grounded to the formulae:

\begin{flushleft}
\begin{itemize}[nolistsep,noitemsep]
\item $\Box ( ( \neg \ocircle observed\_cube\_blue) \lor right\_gripper ) \to  \neg \ocircle pickup\_right ) $ (1)
\item $\Box \Diamond (pickup\_right)$ (2)
\end{itemize}
\end{flushleft}

Formula 1 states that the action proposition \textit{pickup\_right} can not be activated if \textit{observed\_cube\_blue} is \textit{False} and \textit{right\_gripper} is \textit{True}, i.e. the robot cannot pick up if it doesn't see a blue cube or its gripper is full. Formula 2 states the goal of picking up.
The phrase \textit{``Take the red cube''} for the stacking task is grounded to the formulae, which state that the robot cannot pick up with its right gripper if there is no red cube observed or the cube is under a stack, or the gripper is full, together with the goal of picking up with the right gripper:
\begin{flushleft}
\begin{itemize}[nolistsep,noitemsep]
\item $\Box ( ( \neg \ocircle observed\_cube\_red) \lor understack\_cube\_red \lor right\_gripper ) \to  \neg \ocircle pickup\_right )$ (3)
\item $\Box \Diamond (pickup\_right)$ (4)
\end{itemize}
\end{flushleft}

We defined additional features for training the DCG log-linear model which make use of these additional symbols. As in the DCG model, the likelihood of a candidate symbol is computed using \textit{features} that implement specific heuristics depending on the symbols, language and physical environment. The model learns bottom-up symbol structures over a tagged parse tree, such as the one shown in Figure \ref{fig:annotation}. The model defines a \textit{symbol space} containing all symbol types that it can express, such as objects of all possible types and colors and scopes of all possible sensors and actions and objects. 
When training, features are evaluated bottom-up starting from leaf-nodes in the parse tree (i.e. surface words). Features match part-of-speech and symbols at the current level with evidence from lower levels of the tree (both parsed language and symbols), and count as positive or negative evidence in the model, changing the likelihood of symbols corresponding to a sub-tree.
When predicting groundings for a new parse tree, candidate symbols that ground to the parse tree up to that point are evaluated against the language and the underlying symbols, and then the most likely \textit{N} are propagated up the tree. (for all experiments shown in this paper \textit{N=4}).

Thus, first the word terminal contained in the leaf nodes will be associated to a grounding. In the above example, the words ``the box'' are grounded to an \textit{ObjectType} symbol. As the learner moves up in the hierarchy of the tree, each level needs to match or compose symbols from its children. For example, to infer the \textit{Object} grounding of the phrase, the \textit{SpatialRelation} and \textit{Object} symbols must pair accordingly. For the \textit{SpatialRelation} the learner checks if the \textit{Object} coordinates correspond to its type, while for \textit{ObjectType} and \textit{ObjectColor} it matches the object's type and color, respectively. The learning process continues similarly for higher-level symbols; the meaning of the entire instruction is a union of all \textit{Scope} symbols.

We note that the model can be easily extended to ground more complex object references (e.g. nested references) by adding features that interpret such langauge and associate it with symbols \cite{chung2015performance}. 

\section{ENVIRONMENT ASSUMPTION FEEDBACK}\label{sec:assumptions}

Using the previously described symbol hierarchy, our model generates LTL specifications by grounding instructions from the user. We present two complementary contributions for repairing this specification if it is unrealizable: (1) by including formulae describing the initial environment as perceived by the robot; (2) by including liveness formulae describing environment assumptions approved by the user.

To generate a formula describing the environment's initial state, we evaluate all sensor functions on the current environment and add to the specification a formula expressing the environment's initial state. For example, the environment shown in Figure \ref{fig:env2} would have the initial assumptions $(observed\_cube\_blue \land right\_bin\_clear \land observed\_cube\_red \land left\_bin\_clear)$.\\
If the specification is realizable after adding the initial environment formulae, the robot proceeds with executing the synthesized controller and does not prompt the user.


For an unrealizable specification, transitions in the counter-strategy can be used to identify formulae which would limit the environment and allow synthesis \cite{alur2013counter}.
To obtain an environment safety formula, $\phi_e (X_i)$, we use a simplified implementation of this approach, tailored to generating assumptions of the form $\Box \Diamond (\land [\neg] ( X_i ) ),i=1..n$, i.e. a conjunction where ${X_i, i=1..n}$ are sensor propositions that can be negated. This allows us to translate these formulae to binary question prompts for the user.

Formulae (1) and (2) require a single-sensor assumption, $\Box \Diamond observed\_cube\_blue$ , while formulae (3) and (4) require a two-sensor assumptions, $\Box \Diamond observed\_cube\_red \land \neg understack\_cube\_red$.
Our interaction method is not limited to these templates and can be expanded to more complex formulae provided an appropriate prompting method is chosen.

\section{CONVEYING ASSUMPTIONS THROUGH NATURAL LANGUAGE}\label{sec:prompts}

\begin{figure}[b]
	\begin{center}
	\subfigure[]{\includegraphics[width=0.4\linewidth]{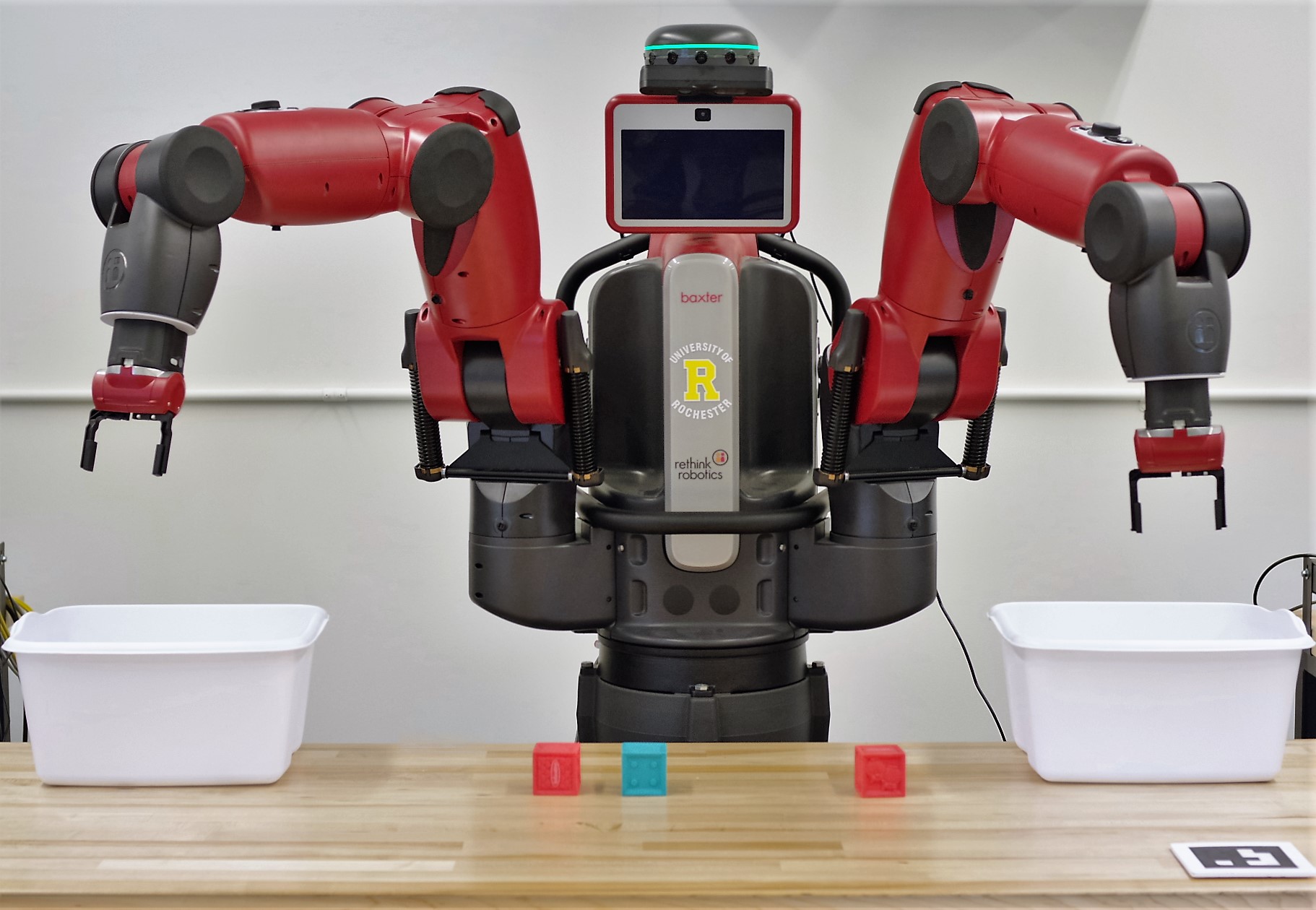}\label{fig:env2}}
\subfigure[]{\includegraphics[width=0.4\linewidth]{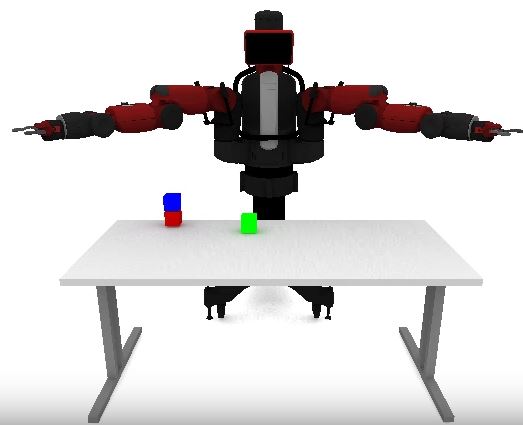}\label{fig:env4}}
	\caption{Physical and simulated grounding environment examples: (a) sorting red and blue cubes, both bins clear; (b) stacking the red cube on the green cube is blocked by the blue cube. \label{fig:environment_examples}}
	\end{center}
\end{figure}

Our design goal for generating prompts was to convey assumptions succinctly in order to minimize the burden on the user. We generate prompts that reference objects in the physical environment, which the user can accept or reject.
We express an environment liveness formula, $\phi_e (X_i)$, where $X_i,i=1..n$ are sensor propositions of type $T_S(i)$ grounded to object $g_o(i)$, through natural language by using a template specific to the sensor type. This template is filled in using the language uttered by the user to reference the object targetted by the sensor proposition. These templates are filled in with an object reference, shown in italics in the following examples, 
which correspond to environments shown in Figure \ref{fig:environment_examples}:
\begin{itemize}[nolistsep,noitemsep]
\item Figure \ref{fig:env2}, one sensor: ``Will \textit{the blue block} remain within reach?''
\item Figure \ref{fig:env4}, two sensors: ``Will \textit{red cube} remain within reach and will you remove \textit{the block on top of red cube}?''
\end{itemize}

\section{EXPERIMENTAL RESULTS}\label{sec:eval}

We evaluated the learning performance our model on two tabletop object manipulation tasks using a Rethink Robotics Baxter Research Robot, for which we designed different specifications in LTL:
\begin{itemize}[nolistsep,noitemsep]
\item \textbf{Cube sorting:} Blue and red were to be sorted into bins placed on either side of the robot; the robot could only pick up a cube if its gripper was empty; the bins could be covered by a lid at any time, preventing cubes from being placed into them (Figures \ref{fig:env2}).

\item \textbf{Cube stacking:} The robot had to pick up a cube of a given color (red, green or blue) and stack it on top of another cube; the robot could only pick up a block if it was not under another cube stacked on top of it (Figure \ref{fig:env4}); similarly, stacking was allowed only if the destination block was not obstructed (Figure \ref{fig:env4}). 

\end{itemize}

We trained a V-DCG grounding model following a procedure similar to \cite{boteanu2016model}: collect NLI using crowdsourcing, annotate a training corpus, and design features for training a model using the corpus.
To collect the training corpus, we executed each specification in eight different simulated environments with varying numbers of blocks and bin behaviors (16 environments in total for the two tasks). For each environment we used manually defined groundings for the propositions, synthesized a controller, and rendered videos of a simulated Baxter executing the task.

We then crowdsourced NLI through surveys\footnote{We deployed identical surveys on http://www.mturk.com for the sorting task, and on http://www.crowdflower.com for the stacking task.}  that showed videos of the robot's execution and required users to freely type instructions that would produce that behavior. The responses consisted of multi-sentence instructions e.g. ``Pick up the innermost blue cube and drop into the right box. Then pick up the remaining blue cube and the innermost red cube. Drop the red cube into the box on the left and then pick up the other red cube. Drop it into the left box and drop the blue one into the right.'' 

From the collected instructions we selected two corpora of multi-sentence instructions: 39 instructions for the sorting task and 21 instructions for the stacking task.
We separated answers into sentences, parsed them using a production grammar and annotated the parsed instructions using the symbol hierarchy shown in Figure \ref{fig:symbols}.
The following are individual training instructions from the annotated training corpus:

\textbf{Sorting:}\\
``Pick up the remaining red cube and drop it in the left bin;''\\
``Pick up the blue block and put it in the box on your right;''\\
``Put the blue cube in the right bin and put the red cube in the left bin.''

\textbf{Stacking:}\\
``Take the red cube. Put the red cube on the green cube;''\\
``Take a green cube. Wait to remove blue cube. Put green cube on red cube.''\\
``Take the blue cube, lift the cube, then place it above the green cube.''

The two-task corpus of 60 instructions (21 stacking, 39 for sorting) contained 828 words (158 unique words) and expressed the following numbers of individual symbols: 153 ObjectType, 109 ObjectColor, 588 Object, 51 SpatialRelation, 60 Region, 275 Sensor, 223 Action, and 222 Scope.
Our model covered (i.e. features could discriminate and represent annotations) 99.9872\% of all the 1,132,710 factors in the joint corpus. 

We evaluated V-DCG on the two-task corpus. Training and testing on all 60 instructions, the model could recover 42 instructions (17 for the stacking task and 25 for the sorting task). Training and testing separately, the model fully inferred 19/21 stacking instructions and 28/39 sorting instructions. Given that the current features offer good coverage of the factor space, as we expand these corpora additional data will improve performance on inferring full instructions.
The annotations between the two tasks have significant lexical, grammatical, and symbolic grounding differences; for example, ``green'' is only used in the stacking task, and ``bin'' is only used in the sorting task. However, the two tasks contain similar phrases referring to red or blue blocks.
The current performance on the joint corpus is slightly lower than individually training models for each task due to an increase in size of the symbol space.

We deployed the trained model on a Rethink Robotics Research Baxter robot and implemented the interaction behavior shown in Figure \ref{fig:interaction} using open-source speech recognition\footnote{http://cmusphinx.sourceforge.net} and text-to-speech\footnote{http://espeak.sourceforge.net/} software.
The robot uses table-top segmentation of RGBD point-clouds to determine properties of the cubes on the table. The location of bins are pre-determined. 
A sampling-based motion planner was used to determine arm trajectories. Grasping of cubes on the final approach is performed by a controller using visual-servoing. Sample instructions are available in the supplemental video\footnote{https://youtu.be/sG8dNGiyNnM}.

\section{CONCLUSION}

In this paper we introduce a natural language interaction method which enables the robot to request the user to approve or reject automatically-generated environment assumptions. The interaction is triggered by synthesis outcomes which identify environment behaviors that would prevent the robot from accomplishing the task. LTL assumption formulae that restrict the environment from exhibiting these behaviors are generated automatically and then expressed through natural language in prompts which the user can accept or reject. 

Grounding complex NLI to the physical environment and LTL formulae enables the synthesis of guaranteed controllers that can accomplish the task described in the instruction. Synthesis guarantees robot behavior under assumptions describing constraints of the robot and the environment. If a grounding model is used in new environments, our contribution can identify new assumptions that will be needed.
\newpage
\bibliographystyle{aaai}
\bibliography{references}

\end{document}